\documentclass{article}
\usepackage[margin=1in]{geometry}
\usepackage{graphicx}
\usepackage{amsmath}
\usepackage{booktabs}
\usepackage{hyperref}
\usepackage[numbers]{natbib}
\usepackage{enumitem}
\usepackage{caption}
\usepackage{titlesec}
\usepackage{placeins}
\usepackage{float}
\titleformat*{\section}{\normalfont\Large\bfseries}
\titleformat*{\subsection}{\normalfont\large\bfseries}
\begin{document}
\begin{center}
 {\LARGE \bf Discrete Tokenization Unlocks Transformers\\[0.4em]for Calibrated Tabular Forecasting}\\[1em]
  {\large Yael S. Elmatad}\\[0.3em]
  {\small \href{mailto:yael.elmatad@gmail.com}{yael.elmatad@gmail.com}}
\end{center}
\vspace{-0.5em}
\noindent\textbf{Code:} \url{https://github.com/yaelelmatad/RunTime-Public}
\section*{Abstract}
Gradient boosting still dominates Transformers~\cite{vaswani2017attention} on tabular benchmarks. Our tokenizer uses a deliberately simplistic discretized vocabulary so we can highlight how even basic tokenization unlocks the power of attention on tabular features, yet it already outperforms tuned gradient boosting when combined with Gaussian smoothing. Our solution discretizes environmental context while smoothing labels with adaptive Gaussians, yielding calibrated PDFs. On 600K entities (5M training examples) we outperform tuned XGBoost~\cite{chen2016xgboost} by 10.8\% (35.94s vs 40.31s median MAE) and achieve KS=0.0045 with the adaptive-\(\sigma\) checkpoint selected to minimize KS rather than median MAE. Ablations confirm architecture matters: losing sequential ordering costs about 2.0\%, dropping the time-delta tokens costs about 1.8\%, and a stratified calibration analysis reveals where miscalibration persists.
\section{Introduction}
XGBoost~\cite{chen2016xgboost} remains the default choice for tabular data because its axis-aligned splits create discrete regimes that Transformers usually miss.  RunTime treats each career as a causal stride of discretized tokens, respecting irregular time deltas so the model learns a full Probability Density Function (PDF) rather than a point estimate. Our approach tackles this by discretizing both inputs and outputs, training with adaptive Gaussian-soft targets, and explicitly representing cadence with time delta tokens. Our contributions are:
\begin{enumerate}[nosep]
  \item Architectural insight that discrete regimes, not bigger Transformers, unlock tabular performance;
  \item Gaussian smoothing - both fixed (\(\sigma\)) and adaptive (\(\sigma_i = \sqrt{\sigma_{\text{floor}}^2 + (k \cdot w_i)^2}\), with \(w_i = b_i^{\text{end}} - b_i^{\text{start}}\)) that scales the smoothing strength with bin width while enforcing a minimum floor;
  \item Empirical win: 10.8\% lower median MAE than tuned XGBoost plus KS=0.0045 calibration (KS optimized via the adaptive-\(\sigma\) best-KS checkpoint);
  \item Analysis methodology including stratified calibration to diagnose residual miscalibration; and
  \item Sequence-aware modeling that explicitly represents cadence with time-delta tokens and relies on entity-disjoint splits to keep temporal dependencies honest for unseen runners.
\end{enumerate}
These design decisions build on prior work demonstrating the effectiveness of discretization for forecasting~\cite{rabanser2020discretization,ansari2024chronos,ansari2025chronos2}, extending the discretized training philosophy from regular time series to irregular, tabular trajectories with entity-level splits and adaptive Gaussian smoothing.
\section{Related Work}
\subsection{Tabular Transformers}
Despite Transformers' flexibility, XGBoost remains dominant on tabular benchmarks. Gorishniy et al.~\cite{gorishniy2021revisiting} show the FT-Transformer loses to XGBoost on 11 datasets, and the TabTransformer~\cite{huang2020tabtransformer} and TabNet~\cite{arik2021tabnet} improve over prior neural methods but stay generally competitive with rather than superior to tree-based models. These models suffer because trees naturally produce piece-wise constant (irregular) decision boundaries through axis-aligned splits, whereas neural networks are inherently smooth function approximators that struggle with the irregular patterns common in tabular data~\cite{grinsztajn2022tree,shwartz2022tabular}. Our work addresses the same failure mode by making discretization explicit and combining it with distributional training.
\subsection{Ordinal Regression and Calibration}
Ordinal regression benefits from soft targets that preserve ordering. Diaz \& Marathe~\cite{diaz2019sord} introduce soft ordinal labels represented as probability vectors over ordinal categories. Calibration literature (Guo et al.~\cite{guo2017calibration}) demonstrates the need for stratified diagnostics beyond global metrics, showing that miscalibration can vary across confidence levels. Kuleshov et al.~\cite{kuleshov2018accurate} develop calibrated regression methods that produce well-calibrated predictive distributions. Our approach integrates Gaussian-integrated targets and stratified calibration during training, rather than relying on post-hoc temperature scaling.

\subsection{Event Trajectories and Temporal Point Processes}
Modeling irregular event trajectories draws on temporal point process foundations (Du et al.~\cite{du2016recurrent}, Shchur et al.~\cite{shchur2020intensity}, Zuo et al.~\cite{zuo2020transformer}), which highlight the need to respect both the event content and the timing gaps. Chronos introduced a similar discretization thesis for regular time series, showing that Transformers trained with discretized horizons match or exceed state-of-the-art forecasting models when each time slot is binned (Ansari et al.~\cite{ansari2024chronos}). Chronos-2 extends that idea toward universal forecasting by mixing discrete value heads with continuous time embeddings (Ansari et al.~\cite{ansari2025chronos2}). Our method extends these insights to heterogeneous bins, irregular histories (requiring explicit time-delta tokens), and entity-disjoint evaluation (preventing individual memorization), bringing the discretization thesis from forecasting into tabular settings rather than strictly generative time-series modeling.

\subsection{Discretization for Forecasting}
Rabanser et al.~\cite{rabanser2020discretization} empirically demonstrate that treating forecasting as classification over discretized bins almost always outperforms regression, attributing the gains to the implicit regularization of binning and the flexibility of distributional outputs. Their method uses hard one-hot targets and assumes uniform bin widths, which ignores ordinal structure and limits handling for wide bins. We extend this line of work by preserving ordinal information through Gaussian-integrated soft targets and by introducing non-adaptive (fixed \(\sigma\)) and adaptive smoothing (\(\sigma_i = \sqrt{\sigma_{\text{floor}}^2 + (k \cdot w_i)^2}\)) to adapt to the heterogeneous target bin widths that arise in binned tabular data.
\section{Method}
\subsection{Problem Setup}
Each training example is a sequence of past races for a single runner. Each race contributes:
\begin{itemize}
  \item environmental conditions (temperature, wind speed, feels-like temperature, humidity, and qualitative conditions such as rain or snow),
  \item race metadata (distance),
  \item demographics (age, gender), and
  \item temporal gaps (\texttt{weeks\_since\_last}, \texttt{weeks\_to\_target}) plus the observed pace.
\end{itemize}
The dataset benefits from the NYRR 9+1 Program qualification history~\cite{nyrr2024}, which ensures a consistent set of marathon and distance event completions per runner, and environmental conditions are sourced from the Visual Crossing Weather API~\cite{visualcrossing2024}.
We discretize pace into 270+ bins and frame the task as predicting the soft distribution across those bins.

\textbf{Entity-disjoint evaluation.} We split 600K runners into train/validation/test (270K/30K/60K runners, ~2.25M/250K/500K examples, 5M total available) with no runner overlap. This prevents the model from memorizing individuals and puts the focus on generalization to unseen runners.
\subsection{Discretization Strategy}
Environmental inputs (temperature, humidity, wind) and the modeled pace are binned via a balanced (quantile-based) quantization procedure so each bin contains roughly the same number of examples, mirroring how tree splits carve regime-specific decision regions. Bins are capped in width, and overly wide extremes are recursively split until their widths stay somewhat comparable to the rest of the vocabulary, keeping the softmax landscape expressive even for rare outcomes. We treat race distance and the two time deltas (\texttt{weeks\_since\_last}, \texttt{weeks\_to\_target}) as tokens directly; they are not quantized via the balanced procedure but are instead encoded from the raw values so the model sees the real timing information. This gives us two token flavors: \textbf{quantized continuous} tokens (pace, weather, etc.) representing discretized numeric ranges, and \textbf{categorical} tokens (gender, weather descriptor, etc.) that encode semantics. Everything in the model—including environmental states, pace bins, demographic cues, and temporal gaps—is represented as a language token, and each event block concatenates the appropriate mix of tokens followed by cadence: \texttt{[features+demographics][pace][d\_next][d\_fin]}. The 327-token window supports up to 30 events and ends with the target pace token to avoid leakage.
\subsection{Gaussian-integrated Soft Targets}
Following Rabanser et al.~\cite{rabanser2020discretization}, we treat the prediction as a classification over discretized bins but replace their hard one-hot targets with Gaussian-integrated soft targets that preserve ordinality:
\[
T_i = \int_{b_i^{\text{start}}}^{b_i^{\text{end}}}
\frac{1}{\sigma\sqrt{2\pi}}\exp\left(-\frac{(x - y_{\text{true}})^2}{2\sigma^2}\right)\,dx,
\]
so bins near the true pace receive consistent credit, preserving ordinal structure.
\subsection{Adaptive Smoothing}
Fixed \(\sigma\) values work well for narrow bins (many middle bins are only 1–3 seconds wide): a 3-second Gaussian produces soft targets for bins of comparable width yet behaves almost one-hot once bin widths significantly exceed the value of \(\sigma\). We therefore scale smoothing via the production quadrature-inspired rule:
\[
\sigma_i = \sqrt{ \sigma_{\text{floor}}^2 + (k \cdot w_i)^2 },
\]
where \(w_i = b_i^{\text{end}} - b_i^{\text{start}}\) is the width of bin \(i\) which is the bin in which the target value lands, \(\sigma_{\text{floor}}\) enforces a minimum smoothing width, and \(k\) controls how aggressively the bin width influences the smoothing. This keeps narrow bins sharp while allowing broad bins to receive proportionally more mass, and the parameter pair \((\sigma_{\text{floor}}, k)\) is tuned on validation data and cached per bin via the training config. The results reported in this paper use \(\sigma_{\text{floor}} = 2.7\) and \(k = 1.5\).
\subsection{Architecture}
A causal Transformer~\cite{vaswani2017attention} (6 layers, 8 heads, 512-dimensional embeddings) processes the token stream. Pace tokens sit before the cadence tokens, and we mask attention to enforce causality. The decoder-style transformer ingests the fixed 11-token stride grammar so every event block is handled like auto-regressive language modeling, producing logits over pace bins that are trained with the Gaussian-smoothed cross-entropy objective described earlier (standard cross-entropy only appears in a specific ablation).

Each event block follows a strict grammar: First the environmental, race specific, and demographic tokens (temperature, humidity, wind, feels-like temperature, rain/clear/other qualitative conditions, gender, age, discretized race distance) then the modeled outcome—pace—sits immediately after the contextual features so the model can learn how environmental/demographic states map into performance, and only then do the two cadence tokens (\texttt{weeks\_since\_last}, \texttt{weeks\_to\_target}) appear to signal the temporal gaps. A block therefore reads \texttt{[race features and demographics][pace][d\_next][d\_fin]}, and at inference time the model autoregressively predicts the final pace token given the preceding context in each stride. Padding tokens fill unused slots so every block keeps the same stride length, which keeps positional encodings (sinusoidal) meaningful even when histories vary from one to thirty events.
\section{Experiments}
\subsection{Setup}
Entity-disjoint splits ensure zero leakage. The test set comprises 60K runners (500K predictions). Evaluation includes MAE, RMSE, and the Kolmogorov-Smirnov statistic for calibration.
\subsection{Results}
Table~\ref{tab:main-results} compares the full RunTime model against three baselines. RunTime achieves a median MAE of 35.94\,s, a 10.8\% improvement over tuned XGBoost (40.31\,s) and roughly 30\% better than the physics-based Riegel formula (49.74\,s). The point-prediction baselines (XGBoost, Naive mean, Riegel) produce identical mean, median, and mode MAE because they emit a single value per example; RunTime's distributional output, by contrast, yields distinct summary statistics---the mode MAE (38.50\,s) is higher than the median (35.94\,s), indicating a slight right skew in the predicted distributions where the most-likely bin is occasionally offset from the probability-weighted center. The RMSE gap is narrower (71.83\,s vs.\ 73.15\,s for XGBoost), which is expected: RMSE penalizes large errors more heavily, and both models struggle on the same long-gap, high-variance examples.
\subsection{Training Configuration}
Table~\ref{tab:train-config} collects the core training settings so the reported convergence behavior can be reproduced at scale.
\begin{table}[h]
\centering
\caption{Main results}
\label{tab:main-results}
\begin{tabular}{lcccc}
\toprule
Model & Mean MAE & Median MAE & Mode MAE & Median RMSE \\
\midrule
Full model (\(\sigma=3\)) & \textbf{36.54} & \textbf{35.94} & \textbf{38.50} & \textbf{71.83} \\
XGBoost (tuned) & 40.31 & 40.31 & 40.31 & 73.15 \\
Naive mean & 52.72 & 52.72 & 52.72 & 88.16 \\
Riegel formula & 49.74 & 49.74 & 49.74 & 94.71 \\
\bottomrule
\end{tabular}
\end{table}

\begin{table*}[h]
\centering
\caption{Training configuration for the fixed \(\sigma=3\) sweep. RunTime is a decoder-style causal Transformer.}
\label{tab:train-config}
\begin{tabular}{ll}
\toprule
Parameter & Value \\
\midrule
Transformer width & 512 \\
Inner feed forward size & 2,048 \\
Dropout after attention/FFN & 0.12 \\
Max races to consider & 30 \\
Attention heads & 8 \\
Decoder-only Transformer layers & 6 \\
Per-GPU batch size & 64 \\
AdamW base learning rate & 1e-4 \\
Gaussian smoothing \(\sigma\) seconds & 3 \\
AdamW weight decay & 0.00118 \\
\bottomrule
\end{tabular}
\end{table*}
\paragraph{Ablation configurations} These MAE statistics come from the dedicated ablation sweep with reduced batch sizes, so they represent slightly more converged checkpoints than the main training runs. We report mean/median/mode to compare the distributional forecasts against tuned XGBoost (median MAE from the sweep) as well as the Naive mean (52.72s) and Riegel (49.74s) baselines, keeping every summary in the same natural units as the ablation runs. The optimizer is AdamW~\cite{loshchilov2019decoupled}, which lets us independently tune weight decay alongside the learning rate for better generalization.
\paragraph{Riegel formula baseline} The Riegel formula extrapolates a runner’s performance across distances by assuming a power-law relation between finishing time and distance~\cite{riegel1981predicting}. We take each runner’s penultimate race as a reference point, normalize its performance by distance, and roll that history forward through the formula to predict the current target pace; this provides a lightweight, physics-inspired benchmark that captures pacing consistency without learning attention weights.
\subsection{Ablations}
\begin{table*}[h]
\centering
\caption{Ablation MAE statistics (\(\sigma=3\)) and wall-clock costs to best MAE.}
\label{tab:ablations}
\begin{tabular}{lrrrr}
\toprule
Configuration & Mean MAE & Median MAE & Mode MAE & Wall clock (h) \\
\midrule
Time token ablation & 37.24 & 36.58 & 39.28 & 107 \\
Time token ablation, shuffled sequence & 37.23 & 36.65 & 39.73 & 145 \\
Full model (\(\sigma=3\)) & \textbf{36.54} & \textbf{35.94} & \textbf{38.50} & \textbf{60} \\
\bottomrule
\end{tabular}
\end{table*}

\begin{figure}[t!]
\centering
\includegraphics[width=1.0\linewidth,height=0.7\textheight,keepaspectratio]{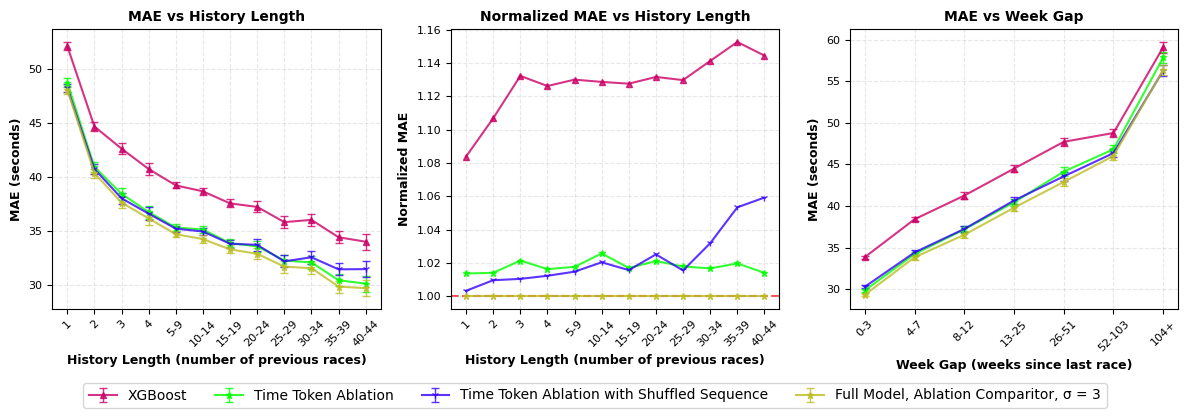}
\caption{Top: MAE vs. history length. Middle: normalized MAE showing the ablation gap relative to the full model. Bottom: MAE vs. week gap between the penultimate and target race, highlighting temporal staleness.}
\label{fig:mae_ablation}
\end{figure}

Table~\ref{tab:ablations} validates that architectural choices—not just model capacity—drive the gains. Figure~\ref{fig:mae_ablation} (middle) shows performance relative to the full model, and the paragraphs below break the contributions down.

\textbf{Time delta tokenization (\(\approx\)1.8\% gain).} Removing explicit time delta tokens raises median MAE from 35.94s to 36.58s (\(\approx\)1.8\%) while the wall-clock to converge jumps from 60h to 107h. This shows that temporal features simultaneously sharpen accuracy and accelerate training.

Even though the accuracy hit remains modest in aggregate, the bottom panel of Figure~\ref{fig:mae_ablation} shows the degradation grows with longer histories, highlighting the cadence tokens’ role in extrapolating far-out predictions. At the same time the full model converges much faster (60h vs 107h) perhaps because the explicit deltas keep the sequence grounded. 

\textbf{Temporal ordering (\(\approx\)2.0\% gain).} Shuffling the race history increases median MAE from 35.94s to 36.65s (\(\approx\)2.0\%), reinforcing that preserving chronological order lets the Transformer capture cadence-dependent progression patterns that random sequence destroys.

\textbf{Temporal generalization.} Figure~\ref{fig:mae_ablation} (bottom) plots MAE vs.\ the week gap between a runner’s penultimate and target race. All variants worsen as the gap grows---closing the gap between the Transformers and XGBoost---suggesting a convergence toward baseline performance at long gaps. The middle panel also shows that the relative gap between the shuffled and full models widens as history length increases, confirming that preserving chronological order helps the Transformer learn progression patterns that generalize better than tree-based methods.
\FloatBarrier
\subsection{Calibration Analysis}
\begin{figure}[t!]
\centering
\includegraphics[width=\linewidth]{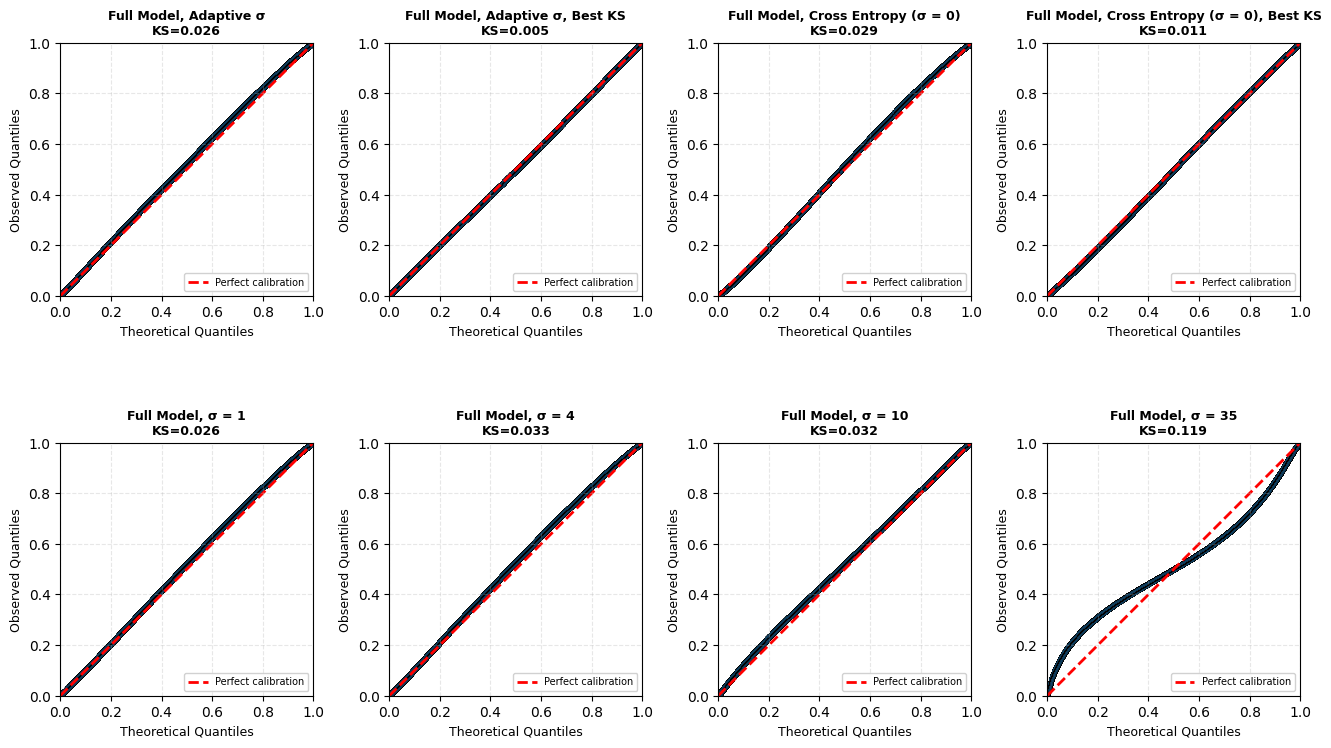}
\caption{Q–Q plot showing distributions across various percentiles from \(\sigma=0\) (cross entropy) to \(\sigma=35\) (wide Gaussian).}
\label{fig:qq-curve-new}
\end{figure}

\begin{figure}[H]
\centering
\includegraphics[width=\linewidth]{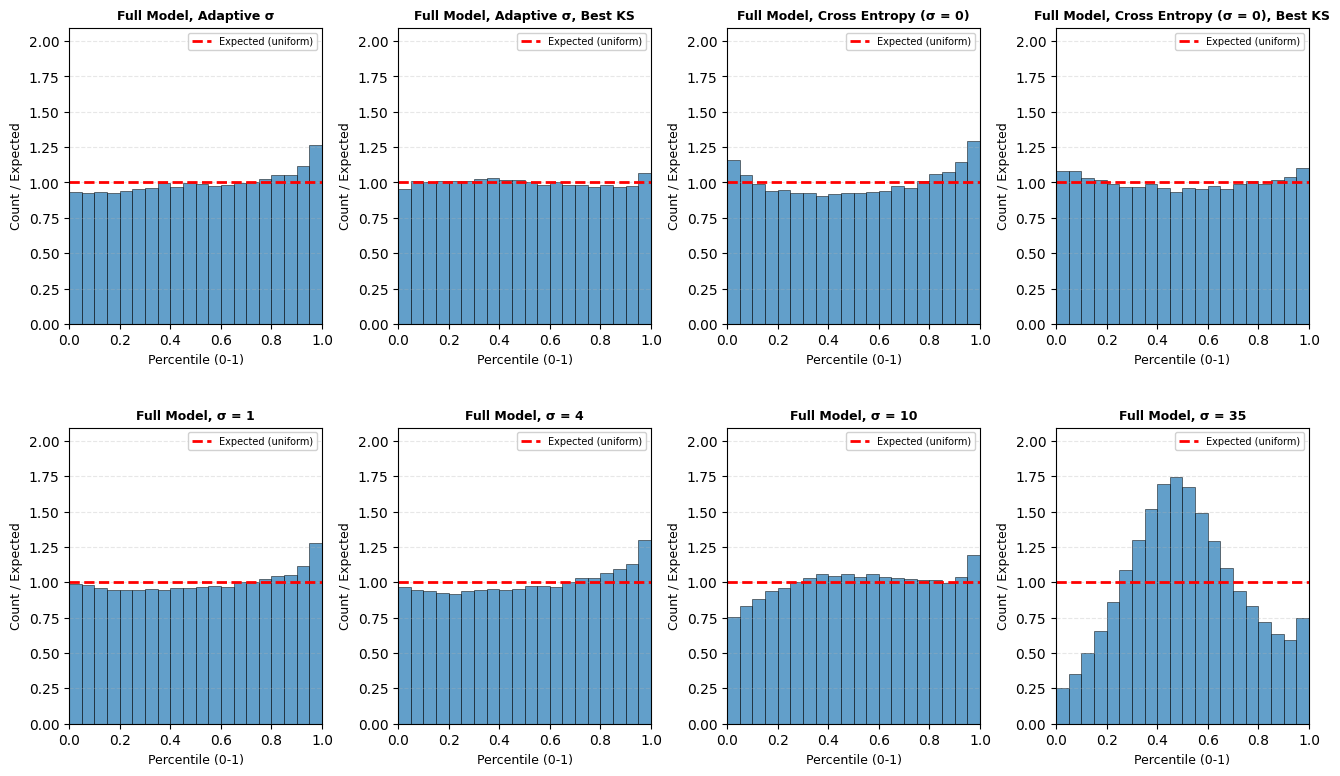}
\caption{Quantile calibration emphasizing where predicted densities concentrate. Note that small \(\sigma\) values concentrate predictions at the extremes, suggesting model overconfidence, whereas large \(\sigma\) values appear underconfident, with larger probability mass in the middle.}
\label{fig:quantile-calibration-new}
\end{figure}

\begin{figure}[t]
\centering
\includegraphics[width=\linewidth]{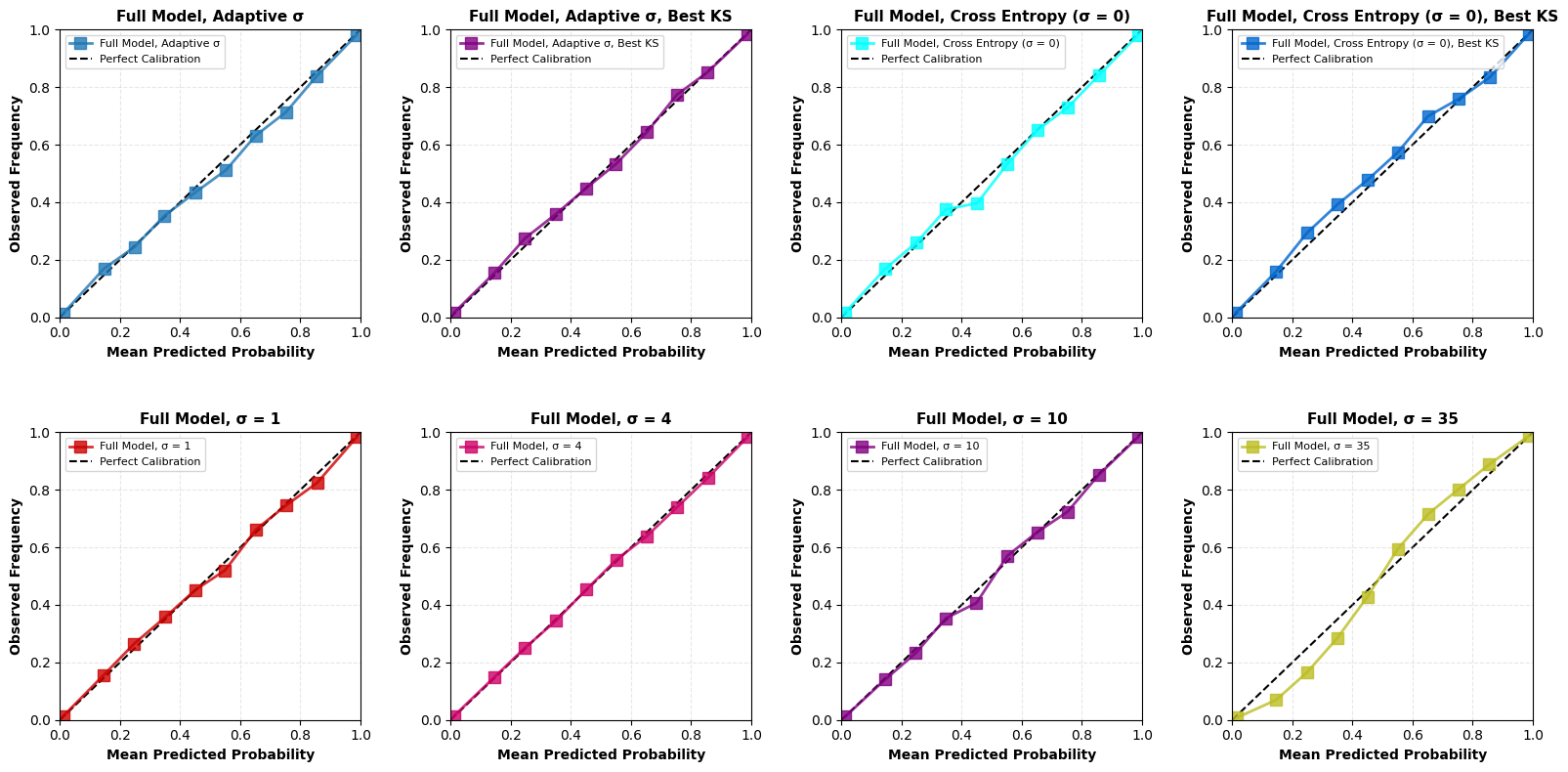}
\caption{Calibration plots highlighting how different smoothing choices affect residual drift across percentiles.}
\label{fig:decile-calibration}
\end{figure}

Figure~\ref{fig:qq-curve-new} shows the updated Q–Q plot staying near uniform (KS=0.0045 from the best-KS adaptive-\(\sigma\) checkpoint), Figure~\ref{fig:quantile-calibration-new} highlights where the predicted probability concentrates its mass, and Figure~\ref{fig:decile-calibration} traces eight decile-specific curves that reveal residual drift for slower runners. These diagnostics demonstrate overall calibration while pointing to the performance-dependent deviations that adaptive smoothing targets, and all three plots were generated using roughly 250,000 evaluation points across 10 bins per decile to keep the comparisons consistent. Importantly this calibration arrives without any post-training temperature scaling—the adaptive \(\sigma\) schedule effectively plays the role of an in-training temperature, and its dependence on bin width and statistics could also motivate a post-training “adaptive temperature” refinement that explicitly compensates for the granular, semi-quantized vocabulary.

\FloatBarrier
\subsection{Calibration Sweep Leaderboard}
Table~\ref{tab:calib-leaderboard} reports the validation MAE/RMSE statistics collected for eight variants of the full model used in the calibration sweep.

\textbf{Important:} these runs used a different training regime than the main result in Table~\ref{tab:main-results}---larger batch sizes, adjusted learning rates, and shorter training schedules designed for fast comparison rather than full convergence. The MAEs in Table~\ref{tab:calib-leaderboard} are therefore systematically higher than the 35.94\,s reported in Table~\ref{tab:main-results} and should only be compared against each other, not against the converged benchmark. The purpose of this sweep is to isolate the effect of the smoothing parameter \(\sigma\) on calibration (KS) while holding the training budget constant.

Rows marked ``Best KS'' refer to the checkpoint selected for the smallest KS statistic, i.e., the calibration-optimized partner for that loss variant rather than the run with the minimum median MAE. The adaptive \(\sigma\) best-KS checkpoint achieves KS\(=\)0.0045 at the cost of slightly higher MAE (37.34\,s vs.\ 36.74\,s for the MAE-optimized adaptive \(\sigma\) checkpoint), confirming that calibration and point accuracy are distinct objectives. All other values of \(\sigma\) had optimized KS values between the adaptive \(\sigma\) sweep and the cross entropy (\(\sigma = 0\)) sweep without improvement in MAE.
\begin{table*}[h]
\centering
\caption{Calibration sweep leaderboard (seconds). Lower is better; values were collected on short, cost-aware sweeps and should only be compared internally.}
\label{tab:calib-leaderboard}
\begin{tabular}{l r rrr rrr}
\toprule
& KS & \multicolumn{3}{c}{MAE} & \multicolumn{3}{c}{RMSE} \\
\cmidrule(lr){2-2}\cmidrule(lr){3-5}\cmidrule(lr){6-8}
Model & Statistic & Mean & Median & Mode & Mean & Median & Mode \\
\midrule
Adaptive \(\sigma\) & 0.0264 & \textbf{36.74} & \textbf{36.22} & \textbf{38.34} & 70.89 & 72.12 & 77.82 \\
Adaptive \(\sigma\), best KS & \textbf{0.0045} & 37.34 & 36.36 & 38.88 & 71.05 & 71.99 & \textbf{76.50} \\
Cross-entropy \(\sigma=0\) & 0.0293 & 37.00 & 36.32 & 38.74 & 71.02 & 72.14 & 77.65 \\
Cross-entropy \(\sigma=0\), best KS & 0.0107 & 37.44 & 36.43 & 39.04 & 71.15 & 72.06 & 77.62 \\
\(\sigma=1\) & 0.0263 & 36.83 & \textbf{36.22} & 38.98 & 70.89 & 72.01 & 82.29 \\
\(\sigma=4\) & 0.0333 & 36.80 & 36.26 & 39.10 & 70.97 & 72.12 & 82.03 \\
\(\sigma=10\) & 0.0321 & 36.80 & 36.23 & 38.94 & \textbf{70.88} & \textbf{71.92} & 80.39 \\
\(\sigma=35\) & 0.1190 & 37.06 & 36.28 & 43.14 & 71.43 & 71.93 & 93.64 \\
\bottomrule
\end{tabular}
\end{table*}

\section{Analysis}
Discretization exposes the discrete regimes trees capture, allowing attention to concentrate within each bin instead of averaging over them. The quadrature-style adaptive smoothing \(\sigma_i = \sqrt{\sigma_{\text{floor}}^2 + (k \cdot w_i)^2}\) keeps the relative smoothing signal consistent by tying it to the target bin width while retaining a floor for narrow bins, which keeps calibration stable even with heterogeneous widths. Treating \(k\) and \(\sigma_{\text{floor}}\) as tunable hyperparameters lets us optimize the adaptive smoothing alongside the other model-selection variables rather than fixing them separately. This adaptive \(\sigma\) also plays a role similar to the temperature parameter used in post-hoc calibration—it rescales the logits, but unlike a post-training reweighting it is learned during training and therefore remains consistent across the vocabulary and bins. MAE decreases from \(\approx48\text{ s}\) at \(h=1\) to \(\approx30\text{ s}\) at \(h=25\), while attention snapshots show the cadence tokens dominate when entropy is high, confirming they supply the temporal context the model needs. Exploring a sigma-reduction schedule (analogous to a learning-rate decay or simulated annealing) might further help the model navigate flatter minima; we plan to test whether progressively sharpening the Gaussian targets during training accelerates convergence without sacrificing calibration.
\section{Discussion and Limitations}
Our experiments confirm that respecting discrete regimes lets Transformers rival tuned gradient boosting. The method generalizes to ordinal regression tasks beyond running; any setting with heterogeneous bin widths can adopt the same discretization plus adaptive smoothing recipe. Limitations include the ongoing adaptive sigma experiments (which we are running), the current lack of overflow bins for tail coverage (which can clip predictions when the pace falls outside the discretized vocabulary), and the fact that all results so far derive from the running dataset—external validation (e.g., MIMIC-IV) remains future work. Adding overflow (and underflow) bins that explicitly model \([b_{\text{max}}, \infty)\) and \((-\infty, b_{\text{min}}]\) would let the model gracefully handle previously unseen extreme paces and is part of the planned extensions.
\section{Conclusion}
Explicit discretization, adaptive Gaussian smoothing, and causal time tokens let a Transformer outperform tuned XGBoost on a large entity-disjoint benchmark while producing better-calibrated PDFs. The paper provides both architectural guidance and analysis tooling (stratified calibration) for future tabular work. One could further treat the adaptive sigma schedule like simulated annealing—starting with larger sigma values and slowly tightening them during training—potentially leveraging the softer targets early on for stability while sharpening distributions as the model converges.
\subsection{Learnable Soft Tokenization}
Our current approach uses fixed quantile binning for the inputs while applying Gaussian soft targets for outputs. A natural extension would apply the same adaptive Gaussian weighting to the input tokenization, creating a symmetric encoding:

\[
\text{embed}(x) = \sum_{i=1}^{K} w_i(x) \cdot \mathbf{e}_i
\]
where
\[
w_i(x) = \frac{\exp\left(-\frac{(x - c_i)^2}{2\sigma_i^2}\right)}{\sum_j \exp\left(-\frac{(x - c_j)^2}{2\sigma_j^2}\right)}
\]

and \(\sigma_i = \sqrt{\sigma_{\text{floor}}^2 + (k \cdot \Delta_i)^2}\) uses the same adaptive smoothing formula from Section~3 (where \(\Delta_i\) is the bin width). This would enable end-to-end learning of optimal bin boundaries while preserving the ordinal structure that discrete regimes provide.
\clearpage
\appendix
\section{Additional Analyses}
\subsection{Runner Trajectories and Calibration Examples}
\begin{figure}[h]
\centering
\includegraphics[width=0.8\linewidth]{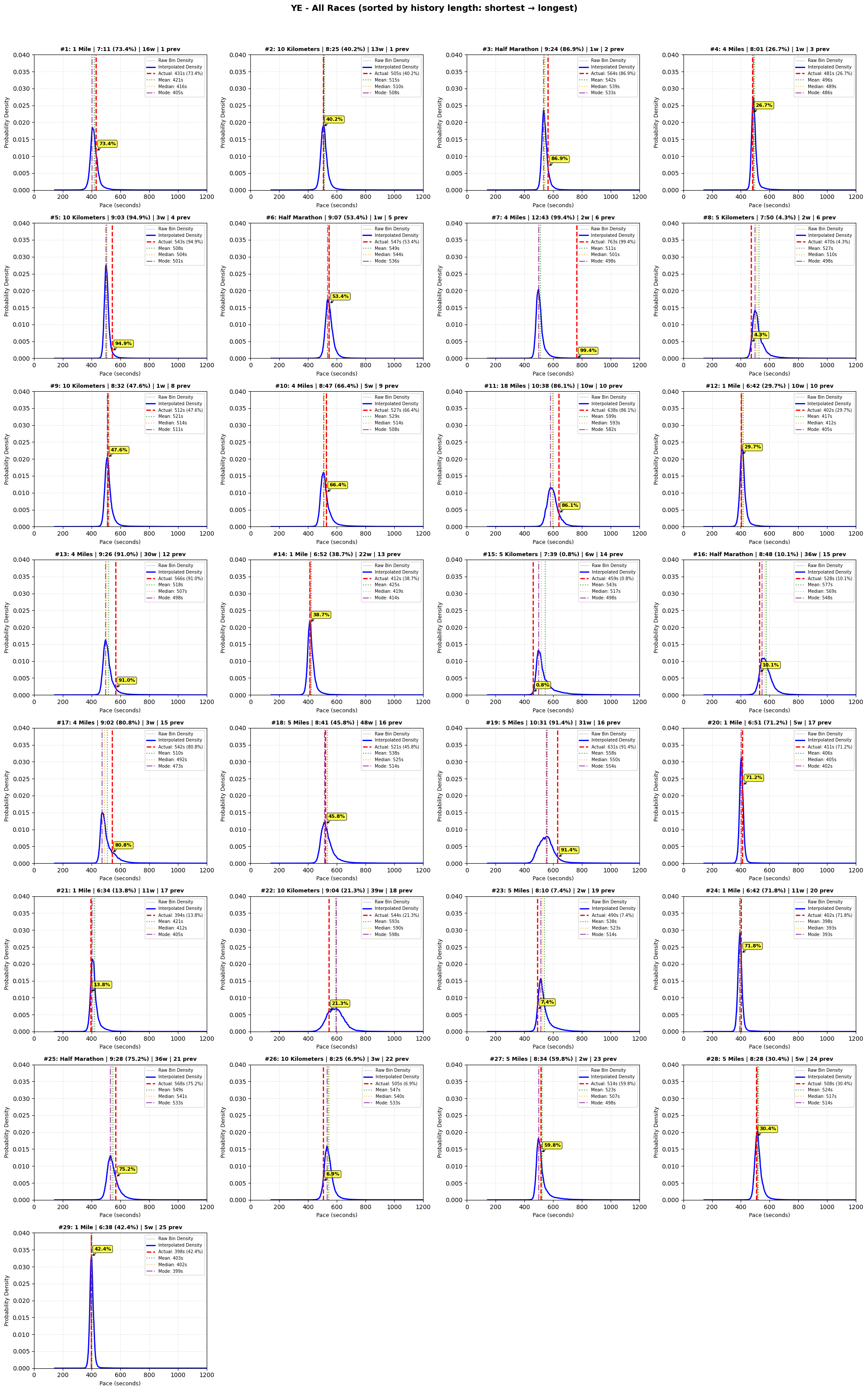}
\caption{YE (short-career) runner trajectory with predicted PDFs illustrating tight uncertainty as the history grows.}
\label{fig:runner-ye}
\end{figure}

\begin{figure}[h]
\centering
\includegraphics[width=0.9\linewidth]{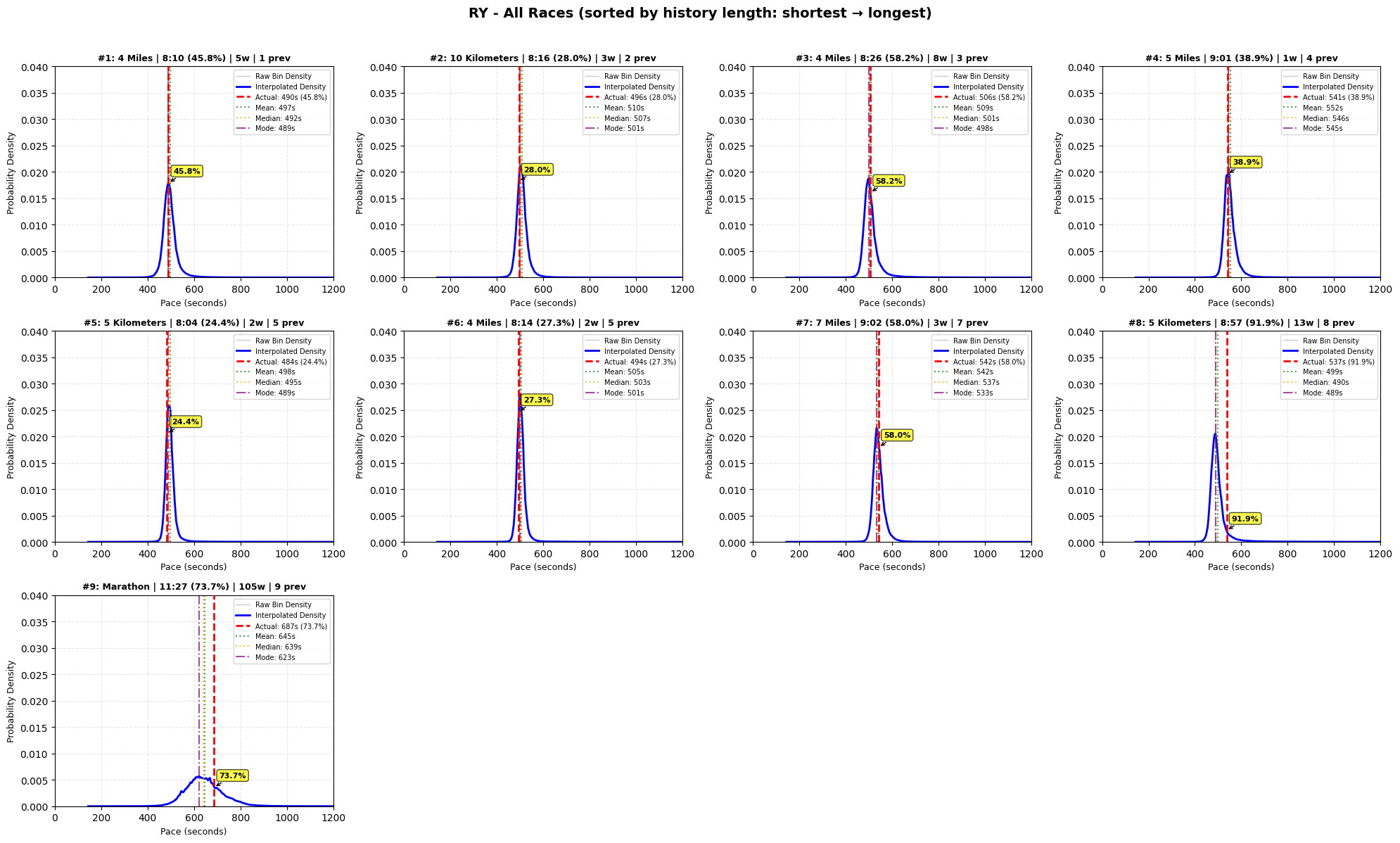}
\caption{RY (median-career) runner trajectory showing moderate calibration and widening distributions for longer distances.}
\label{fig:runner-ry}
\end{figure}

\begin{figure}[h]
\centering
\includegraphics[width=0.9\linewidth]{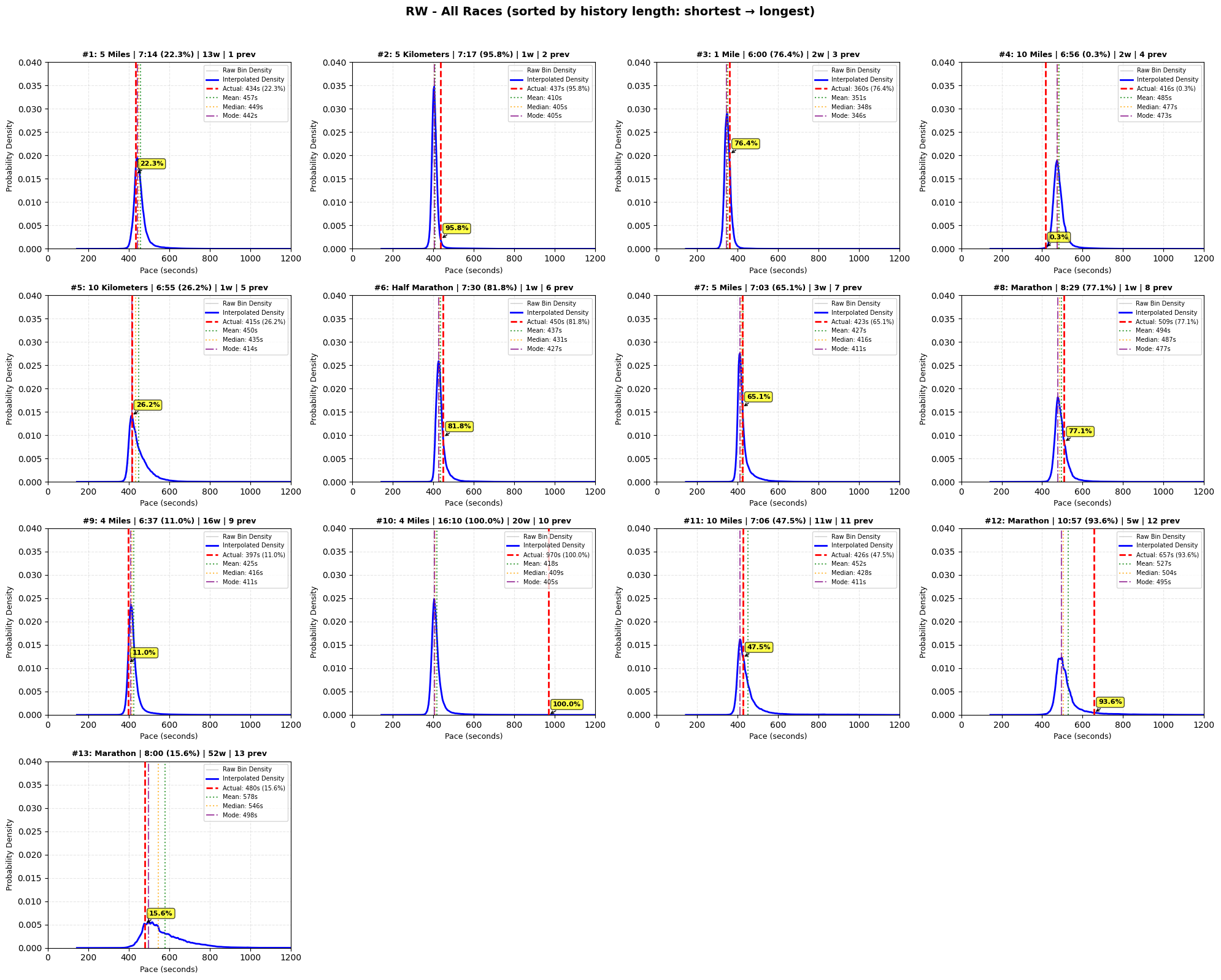}
\caption{RW (long-career) runner trajectory that exhibits sustained variability and broad PDFs even late in the history.}
\label{fig:runner-rw}
\end{figure}

Figure~\ref{fig:runner-ye} through Figure~\ref{fig:runner-rw} show three representative runner histories (short, median, and long careers). Each block overlays the observed pace sequence with the Gaussian-smoothed prediction envelopes, illustrating the full predicted distribution (PDF) across the vocabulary and how the density sharpens for consistent performers while broadening for more erratic trajectories.

\paragraph{Runner selection note} These three runners were selected because ground-truth context is available for their outlier races, which lets us verify that prediction errors reflect human decision-making rather than modeling failure. For example, Runner~A's slower-than-usual race~\#7 and race~\#19 correspond to known non-competitive efforts (pacing a companion). The point is to show that even when the model is well-calibrated, humans still choose to vary their strategy, so distributional forecasts are the right interface for uncertainty-aware guidance.

\subsection{Activation and Attention Diagnostics}
\begin{figure}[h]
\centering
\includegraphics[width=0.95\linewidth]{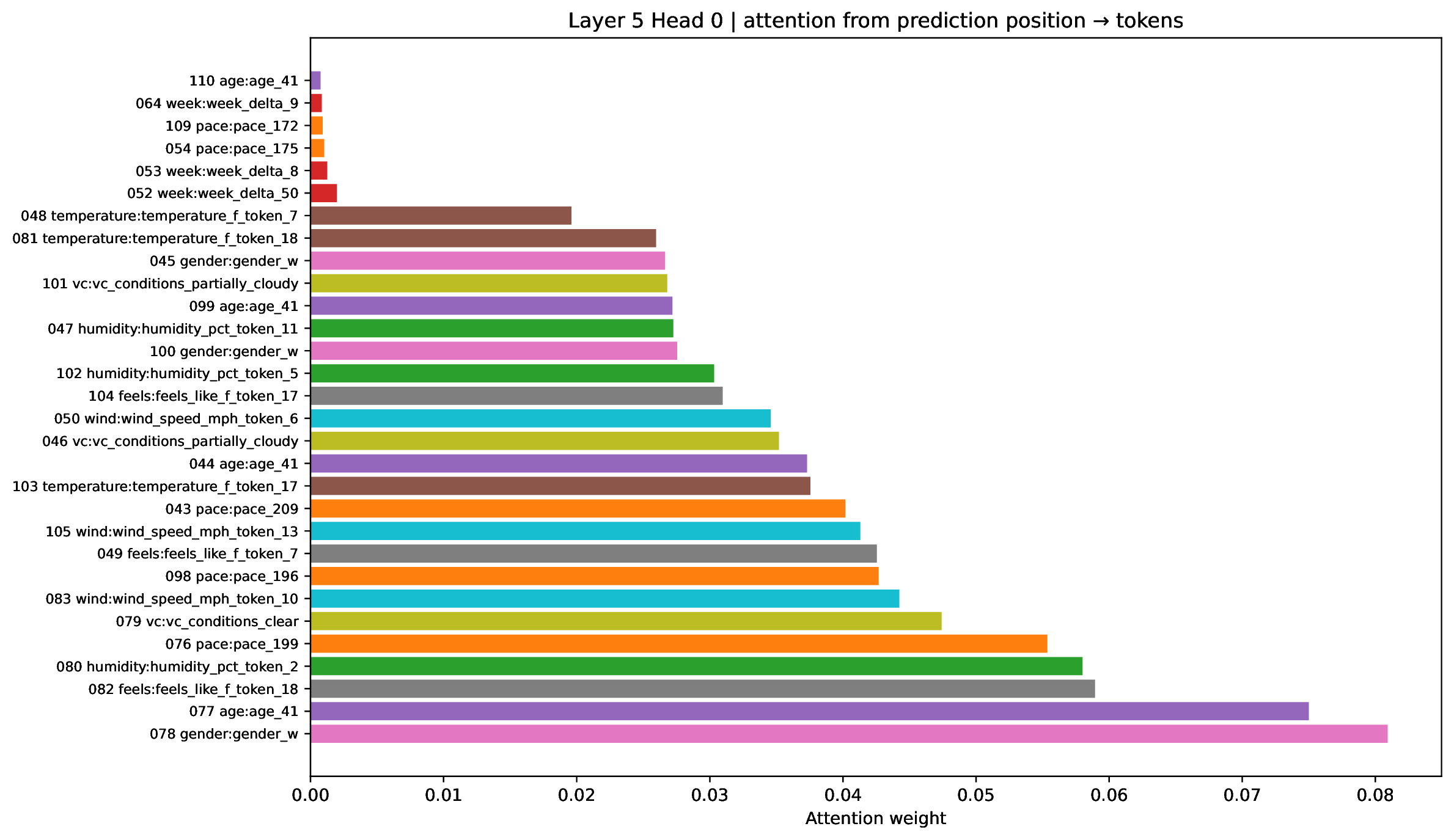}\\[0.3em]
\includegraphics[width=0.95\linewidth]{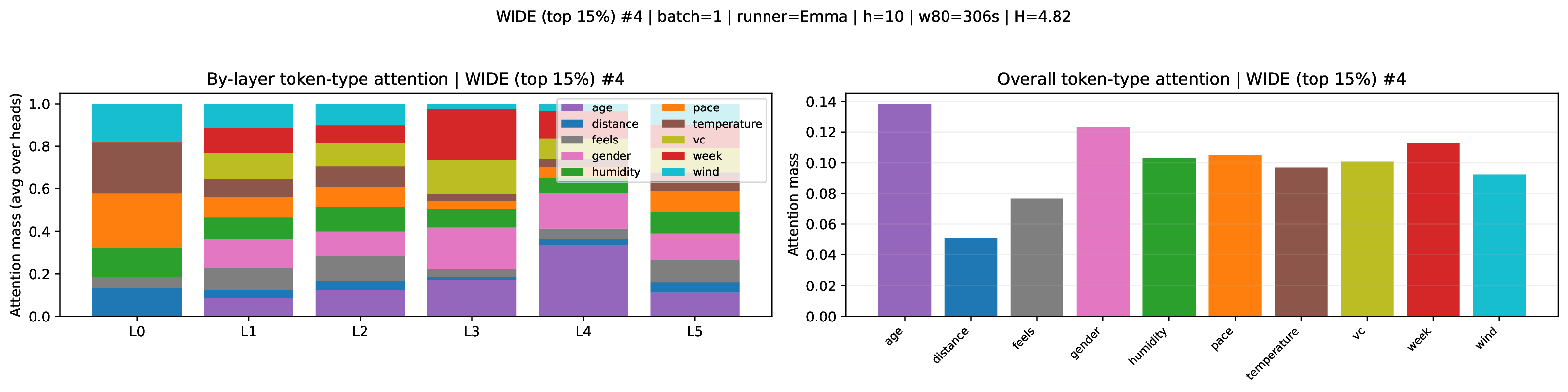}\\[0.3em]
\includegraphics[width=0.95\linewidth]{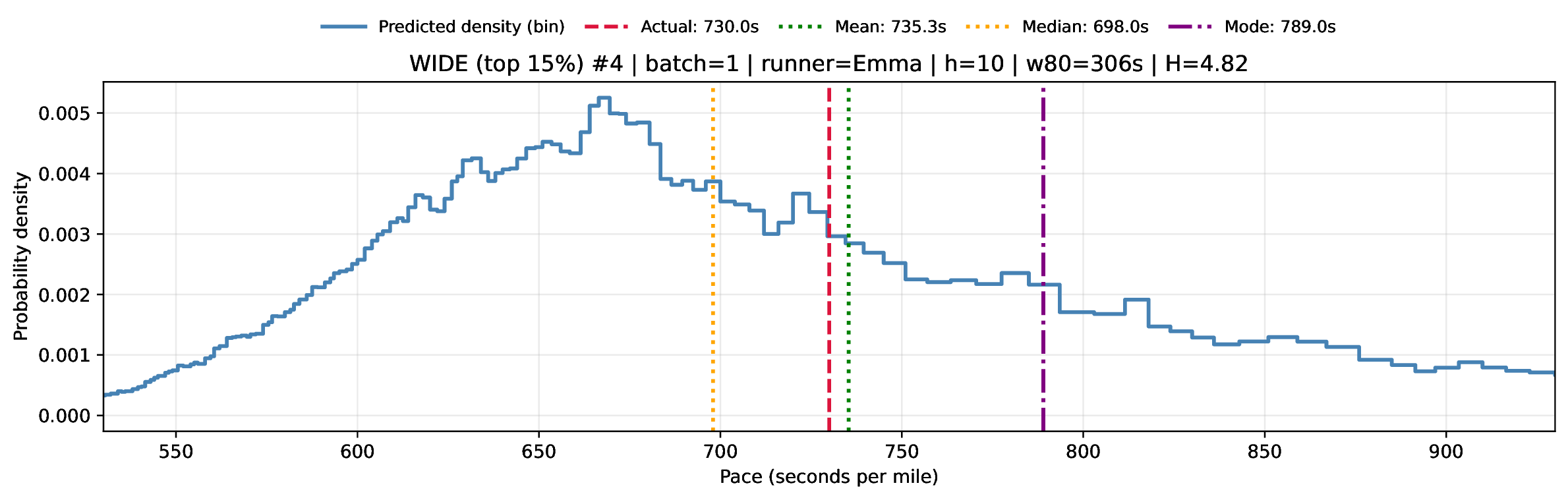}
\caption{Activation and attention diagnostics for a high-uncertainty prediction. Top: attention weights for the final transformer block. Middle: per-layer contribution magnitudes. Bottom: pace distribution histogram showing the predicted softmax.}
\label{fig:activation-visualization}
\end{figure}

Activation statistics and attention patterns confirm that cadence tokens dominate when entropy is high, while the pace bins sharpen when variance is low. Figure~\ref{fig:activation-visualization} visualizes the same batch from three views (attention, per-layer contributions, and distributional histogram), showing how the model routes focus toward time deltas during uncertain predictions.

\subsection{Data Provenance and Survival-Analysis Context}
The core dataset is drawn from the NYRR publicly available results, with the NYRR 9+1 Program ensuring a large, consistent set of races for many athletes~\cite{nyrr2024}, and the environmental tokens rely on Visual Crossing’s historical weather API~\cite{visualcrossing2024}.  Adaptive sigma caches follow the implementation in the training code, matching the quadrature-style smoothing and \((\sigma_{\text{floor}}, k)\) pairs described earlier.

\subsection{Grammar Efficiency Considerations}
The current stride grammar repeats entity-level signals such as gender or baseline demographics in every block, which mirrors how the data is stored but adds redundant tokens that could be compressed. A more concise alternative would separate the inherent runner features (fixed across the career) from the event-specific descriptors (distance, weather, pace) and the time tokens, only reintroducing the moving parts at each step. Doing so would preserve the causal story (fixed features → event features → timing → next event features) while saving memory and speeding up training, since fewer repeated embeddings would need to be cached. We keep the redundant copies for transparency and because they simplify batching, but future grammar designs could exploit this structure to cut down on per-stride verbosity.

\subsection{Future Work and Applications}
Several natural extensions flow from the discretized grammar. One strand treats the pace PDF as a survival-style signal: adding censoring tokens such as \texttt{time\_censored}, \texttt{time\_eos}, and \texttt{time\_missing} can flag trajectories where no future race is observed within the window, so the Transformer can distinguish “still at risk” careers from ones that terminated. This mirrors classic survival analysis~\cite{cox1972regression,katzman2018deepsurv,lee2018deephit} while preserving the irregular attention-based representation; mechanisms like right-censoring or overflow bins (e.g., explicit bins for \(y < b_1^{start}\) or \(y > b_K^{end}\)) let the PDF gracefully extend beyond the observed support.

Another strand reinterprets the causal stride as a generative process. A pure generative Transformer can sequentially sample every token within an event block, conditioning on the previous context and respecting hybrid loss functions that treat discrete tokens (categorical weather) with cross-entropy and quantized continuous tokens (pace, numeric environmental features) with smoothed Gaussian penalties. The specific token ordering becomes a hidden hyperparameter in this setup: choosing which environmental covariates to generate first before the outcome changes which conditional factors the model learns, and one could pursue both conditional generation (treating future covariates as given) and full joint generation (ordering tokens to match a plausible data-generating story).

Such an auto-regressive sampler enables Monte Carlo digital twins: recursively sampling from the predicted PDF at every step produces families of plausible future trajectories, which support stress testing, tail-risk analysis, or scenario planning in domains like energy, flood management, or finance. These trajectories can also be conditioned on arbitrary covariate sequences to simulate missing data (e.g., inserting sentinel tokens for the absence of temperature or opponent identity) or to explore heterogeneous event grammars (race vs training run events). Together, these extensions underscore how the same discretized architecture scales from distributional forecasting to generative simulation while retaining calibration and interpretability.
 
\subsection{Huber-loss exploration}
An early experiment combined the Transformer’s softmax head with a Huber-style point-loss on the expected pace as an auxiliary objective. The idea was to force the model to hedge around the prediction while still learning a distribution; the result was instructive. Under this composite objective the model learned to route almost all probability mass to the extreme pace bins (first and last of the vocabulary) while adjusting their relative weights to hit whatever point estimate the Huber loss demanded. In other words, the model “cheated” the loss by ignoring the interior bins and collapsing the distribution onto the extremes, so the apparent MAE improved at the cost of losing the predictive PDF entirely. This failure reinforced the lesson that RunTime must prioritize distributional fidelity—hence the switch to full Gaussian-smoothing with cross-entropy as described in the main text.
\clearpage
\bibliographystyle{unsrtnat}
\bibliography{references}
\end{document}